\documentclass[accept]{article}
\usepackage{ijcai23}

\usepackage{times}
\usepackage{soul}
\usepackage{url}
\usepackage[hidelinks]{hyperref}
\usepackage[utf8]{inputenc}
\usepackage[small]{caption}
\usepackage{graphicx}
\usepackage{amsmath}
\usepackage{amsthm}
\usepackage{booktabs}
\usepackage{algorithm}
\usepackage{algorithmic}
\usepackage[switch]{lineno}
\usepackage{multirow}
\usepackage{subfigure}
\usepackage{tablefootnote}
\title{Chinese Fine-Grained Financial Sentiment Analysis with Large Language Models}

\author{
Yinyu Lan
\and
Yanru Wu\thanks{Corresponding Author}\and
Wang Xu \And 
Weiqiang Feng\And
Youhao Zhang
\affiliations
FinChina AI Research\\
\emails
lanyinyu19@mails.ucas.ac.cn,\{wuyr, xuwang, fengwq, zhangyh\}@finchina.com
}

\begin{document}

\maketitle

\begin{abstract}

Entity-level fine-grained sentiment analysis in the financial domain is a crucial subtask of sentiment analysis and currently faces numerous challenges. The primary challenge stems from the lack of high-quality and large-scale annotated corpora specifically designed for financial text sentiment analysis, which in turn limits the availability of data necessary for developing effective text processing techniques. Recent advancements in large language models (LLMs) have yielded remarkable performance in natural language processing tasks, primarily centered around language pattern matching. In this paper, we propose a novel and extensive Chinese fine-grained financial sentiment analysis dataset, FinChina SA, for enterprise early warning. We thoroughly evaluate and experiment with well-known existing open-source LLMs using our dataset. We firmly believe that our dataset will serve as a valuable resource to advance the exploration of real-world financial sentiment analysis tasks, which should be the focus of future research.\footnote{The FinChina SA dataset is publicly available at \url{https://github.com/YerayL/FinChina-SA}.}

%Our dataset and all code to replicate the experimental results will be released.

\end{abstract}

\section{Introduction}
\begin{figure}[!t]\centering
	\includegraphics[width=9cm]{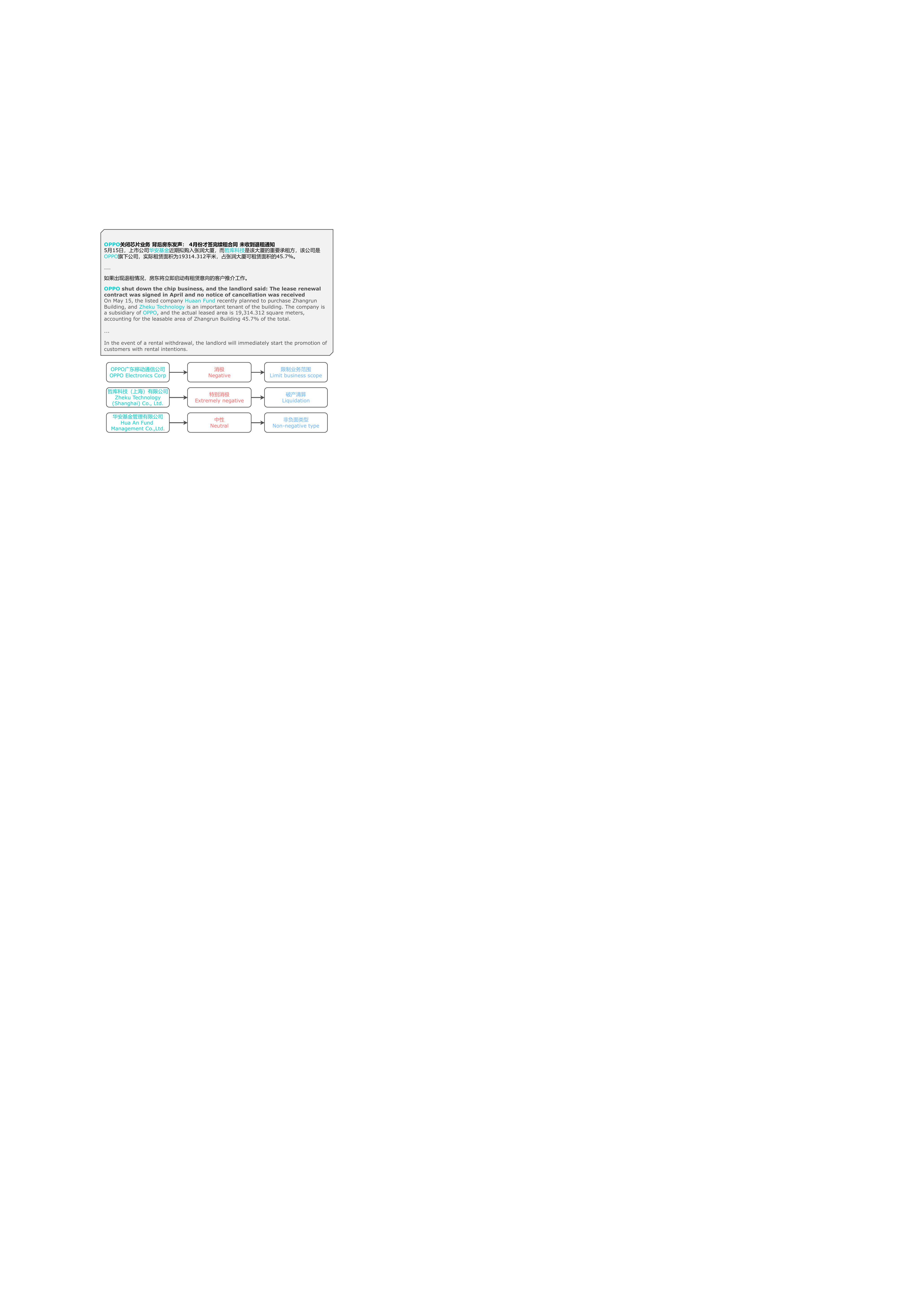}
	\caption{An example of financial sentiment analysis task for enterprise early warning. The cyan font indicates the companies involved in the news. The red and blue fonts indicate the associated polarity and warning type respectively.}
    \label{FIG_1}
\end{figure}

The rapid development of the Internet and the financial industry has led to an abundance of professional stock review reports, research reports, and individual investors' opinions and analyses in the financial domain. Whether in news reports or commentaries on related topics and companies, these texts often provide evaluations and attitudes towards relevant events and companies, offering valuable insights for investment and regulation. A comprehensive understanding of this evaluation information can enhance investors' market understanding and aid in investment decision-making. Additionally, enterprises or financial market regulators can benefit from early identification of hidden issues through such evaluations, enabling them to grasp market dynamics and mitigate risks. Consequently, sentiment analysis of financial texts has emerged as a prominent area of research and application.  The text entity-level fine-grained sentiment analysis research in the financial domain is still in its infancy, and it is also an important subtask of fine-grained sentiment analysis \cite{pang2008opinion}, and currently facing many challenges. However, there is a lack of high-quality and large-scale text annotation corpus in the Chinese financial domain, resulting in a lack of data support for the underlying technology. 

Large language models have attracted a lot of attention in the domain of artificial intelligence due to their impressive ability to solve natural language processing tasks. In particular, dialogue-based large language models such as ChatGPT have had a major impact on social development and have played a crucial role for the application of artificial intelligence in daily life, attracting widespread attention from academia and industry. However, their research in specific domain, such as the financial domain, is still relatively scarce. How to better use and evaluate the ability of large language models in the vertical domain can solve the research problems that need to be solved urgently \cite{li2023chatgpt}.

Since the release of ChatGPT, there has been a growing number of related models developed and released, based on LLaMA \cite{llama} and BLOOM \cite{bloom} models. Recent research has predominantly focused on effectively leveraging general LLMs for domain-specific supervised fine-tuning training. This process entails constructing a dataset in either instruction or conversational format and fine-tuning it using a pre-trained backbone model. Notably, this way of training with significantly smaller data than used in pre-training can yield favorable results. These works can be categorized into two distinct types \cite{chen2023phoenix}. The first category is instruction-based tuning, with Alpaca \cite{alpaca} serving as a prominent example. Alpaca utilizes self-instruction \cite{self-instruct} techniques to generate supplementary instructions for fine-tuning the GPT 3.5 model, leading to enhanced accuracy and context-sensitive output. The second category comprises conversation-based tuning models that leverage the distillation of user interactions with ChatGPT. Vicuna \cite{vicuna2023} exemplifies this approach by utilizing a comprehensive dataset of user-shared ChatGPT conversations to improve model performance.

Aiming to address the lack of a comprehensive Chinese financial sentiment analysis dataset and meet the demands of enterprises regarding negative news alerts, we propose the FinChina SA dataset specifically designed for the financial domain. Figure \ref{FIG_1} illustrates an example of the financial sentiment analysis task for enterprise early warning. Initially, we conduct data crawling and cleaning on major financial news websites. Subsequently, We label the company names, sentiment polarities and warning types in the crawled data. A total of 11,036 news articles are annotated, including 8,739 companies, 190 warning types, and 21,272 corresponding sentiment examples. To explore the application of LLMs in the financial domain, we conduct extensive research and experiments on the FinChina SA dataset. We train these models using instruction-based tuning and conversation-based tuning techniques. The experimental results demonstrate the promising performance of financial LLMs (FinLLMs) and the limitation of ChatGPT in financial sentiment analysis tasks.

The contributions of this paper are as follows: (1) We propose FinChina SA, a novel large-scale dataset designed for fine-grained sentiment analysis in the Chinese financial domain. (2) We evaluate and analyze the capabilities and limitations of ChatGPT in the zero-shot setting for fine-grained financial sentiment analysis. (3) We compare and analyze the capabilities of well-known existing open-source LLMs on the benchmark and discuss the feasibility and prospects of developing FinLLMs.

\section{Related Work}

Following a comprehensive examination of research pertaining to fine-grained financial sentiment analysis (Section 2.1) and large language models (Section 2.2), we delve into the discussion of supervised fine-tuning of LLMs (Section 2.3).

\subsection{Fine-Grained Financial Sentiment Analysis}
Fine-grained sentiment analysis is a significant task with substantial practical applications. However, there is a scarcity of studies focused on entity-level fine-grained sentiment analysis within the financial domain, particularly due to the lack of available datasets. This limitation poses challenges for conducting research in this task. \cite{Semeval2017} analyzed the approaches and tools employed by over 30 participants during the SemEval-2017 conference on "Sentiment Analysis of Financial Microblog and News." The majority of participants relied on traditional machine learning models such as SVM and SVR. \cite{do2019deep} highlighted the demanding nature of data labeling within the financial domain, emphasizing the need for domain expertise and the associated high costs. Consequently, limited labeled data is available. \cite{www2018open} publicly released a small dataset named FiQA, comprising textual instances within the financial domain. The dataset also includes the entities referenced in the text, with corresponding sentiment scores assigned to each entity.

\subsection{Large Language Models}
The introduction of the transformer model \cite{transformer} has enabled the training of unsupervised text data at a large scale. Over the past few years, encoder-based models, such as BERT \cite{devlin2018bert}, have exhibited impressive capabilities in various natural language processing (NLP) tasks. More recently, decoder-based models, including GPT-1 \cite{gpt1}, GPT-2 \cite{gpt2}, and T5 \cite{T5}, have made significant advancements. With the increasing number of model parameters, models like GPT-3 \cite{gpt3}, often referred to as LLMs, have gradually acquired zero-shot learning abilities. These models can generate responses based on instructions without relying on examples. Moreover, \cite{wu2023bloomberggpt} have recently proposed a proprietary model, BloombergGPT, in the financial domain, focusing primarily on pre-training. An open-source large language model, FinGPT, is proposed by \cite{yang2023fingpt}. It takes a data-centric approach, providing researchers and practitioners with accessible and transparent resources to develop their FinLLMs. 

\subsection{Supervised Fine-Tuning}
Currently, post-training methods that are used in supervised fine-tuning of LLMs can be categorized into instruction-based tuning and conversation-based tuning \cite{chen2023phoenix}. Instruction tuning aims to train language models to comply with human instructions \cite{instrucGPT}, which can be manually designed or created in a hybrid manner where humans provide initial instructions and OpenAI ChatGPT generates additional similar instructions using in-context learning \cite{self-instruct}. Language models are taught to engage in chat-based conversations similar to OpenAI ChatGPT through the use of ChatGPT-distilled conversations, whereas instruction data is typically employed for single-turn question answering (QA). A notable example of this approach is Alpaca \cite{alpaca}, which utilizes the self-instruction technique to generate additional instructions using the GPT 3.5 model for fine-tuning. This leads to improved accuracy and contextually relevant outputs. The second category consists of conversation-based tuning models that leverage the distillation of user interactions with ChatGPT. Vicuna \cite{vicuna2023} is an exemplary model in this category, benefiting from extensive user-shared dialogue datasets to enhance model performance.

\section{Data Collection}
\iffalse
\begin{figure}[!t]
\centering
\subfigure[]{
\begin{minipage}[t]{0.3\linewidth}
\centering
	\includegraphics[scale=0.3]{distribution.pdf}
 \end{minipage}%
}%
\hspace{0.7in}
\subfigure[]{
\begin{minipage}[t]{0.4\linewidth}
\centering
\includegraphics[scale=0.3]{examples.pdf}
%\caption{fig2}
\end{minipage}%
}%
\centering
\caption{Distribution of news text lengths (\textbf{a}) and the number of negative sentiments and non-negative sentiments (\textbf{b}).}
\label{FIG_2}
\end{figure}
\fi

\begin{figure}[!t]\centering
	\includegraphics[width=7cm]{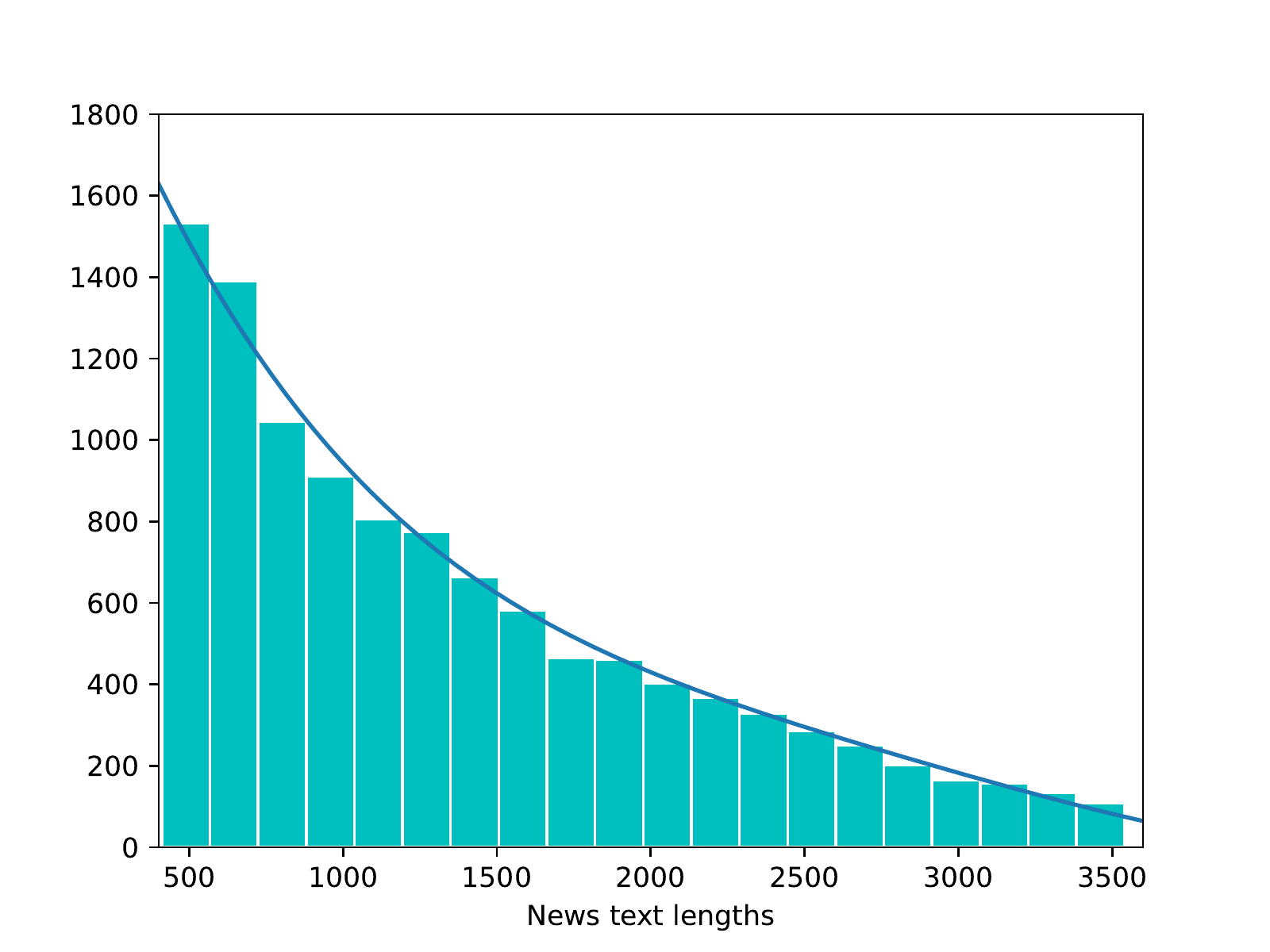}
	\caption{Distribution of news text lengths. }
    \label{FIG_2}
\end{figure}

\begin{figure*}[!t]\centering
	\includegraphics[width=18cm]{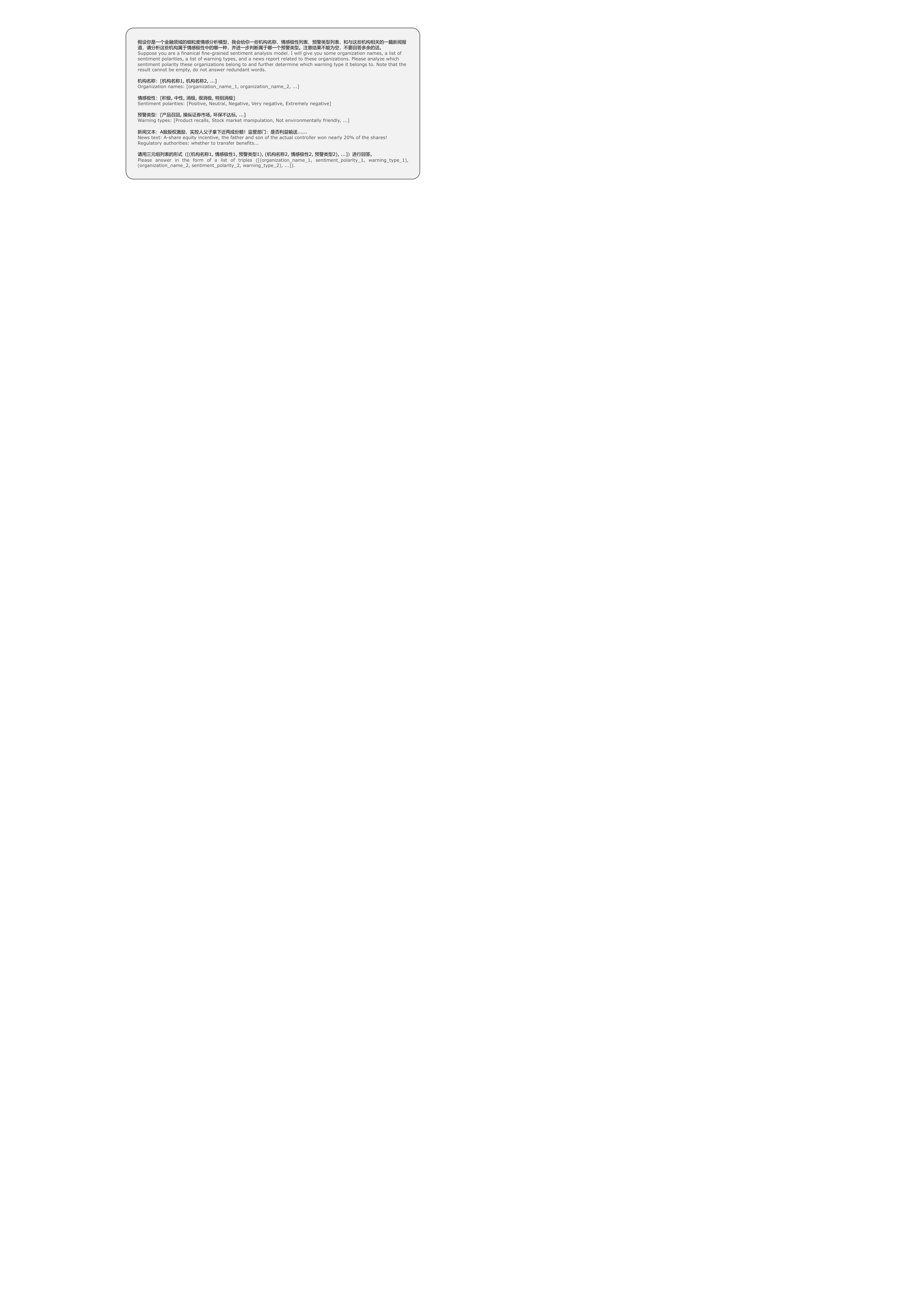}
	\caption{Illustration of the prompt template used in our experiments for ChatGPT}
    \label{FIG_3}
\end{figure*}

In order to construct a comprehensive sentiment analysis dataset in the financial domain, we recruited a team of 20 annotators specializing in finance-related subjects. Initially, we selected 11,036 representative high-quality news articles from a pool of 288,788 crawled news articles. These articles served as the basis for annotation. Each annotator pre-labeled 2,500 data samples, and we compared the respective labeling results during the annotation process. We documented the differences and ambiguities and used them to formulate comprehensive labeling guidelines for various entities and contexts that exhibited ambiguity or conflict. During the labeling phase, each news text was independently reviewed by a minimum of 5 annotators. It is worth mentioning that the annotators did not engage in communication during the labeling process and solely relied on the established guidelines. Once the independent labeling was completed, any discrepancies or errors in the labeling results were addressed through discussions involving an additional annotator. The aim was to reach a consensus among all annotators, leading to modifications in the labeled data and the final completion of the annotation process.

Out of the 21,272 companies involved, 6,991 are positive sentiments, while 14,281 are negative sentiments. The specific statistical findings are presented in Table \ref{tab1} and Figure \ref{FIG_2}. Given that our sentiment analysis primarily focuses on enterprise early warning, we place particular emphasis on negative emotions. To further categorize negative emotions, we divide them into three levels: extremely negative, very negative, and negative. Among the negative sentiments, there were 225 instances of extreme negativity, 1,886 instances of high negativity, and 1,170 instances of general negativity.

\begin{table}[!t]
\centering
\caption{Statistics of the FinChina SA dataset}
\label{tab1}
\setlength{\tabcolsep}{4mm}
\begin{tabular}{ll}
\hline\noalign{\smallskip}
\bfseries Stats &  \bfseries Number \\
\noalign{\smallskip}\hline\noalign{\smallskip}
\# News texts & 11,036\\
\# Companies & 8,739\\
\# Sentiments & 21,272\\
\# Non-negative sentiments & 6,991\\
\# Negative sentiments & 14,281\\
\# Early warning types & 190\\
Max news text length & 3,540\\
Min news text length & 411\\
Avg. news text length & 1,357\\
\noalign{\smallskip}\hline
\end{tabular}
\end{table}

\section{Models}

Sections 4.1-4.5 introduce the pre-trained language models used in our experiments in detail, including models such as Longformer, LLaMA, BLOOM, ChatGLM and ChatGPT.

\subsection{Longformer}
To compare with previous pre-trained models having fewer parameters, we utilize the Longformer model \cite{Beltagy2020Longformer}. The Longformer can process sequences up to 4,096 tokens by employing an attention mechanism that scales linearly with the input text length, unlike the quadratic behavior in earlier Transformer models such as BERT \cite{devlin2018bert}. Considering the maximum document length in our corpus (over 3,540 words, see Table \ref{tab1}), we deemed the Longformer model a suitable baseline. The Longformer model is used to process the initial 4,096 tokens of each document, and the resulting 768-dimensional pooled output is retained as the document representation. This representation is then fed into two feed-forward classification neural network to predict probabilities for all sentiment polarities and warning types. The Longformer weights undergo additional fine-tuning during the training process for our classification task. Since our dataset is in Chinese, we use the Longformer-chinese\footnote{\url{https://github.com/SCHENLIU/longformer-chinese}} which is pre-trained on Chinese news corpus.

\subsection{LLaMA}
LLaMA \cite{llama} is a collection of open-source multi-lingual base models with parameter sizes ranging from 7 billion to 65 billion, made available to the research community. However, the original LLaMA-7B model lacked Chinese corpus during pre-training, resulting in the absence of Chinese vocabulary. Consequently, we employed the Chinese LLaMA model \cite{chinese-llama-alpaca}, which incorporated an expanded Chinese vocabulary, and conducted a secondary pre-training using the 120G Chinese general corpus known as Chinese LLaMA plus \cite{chinese-llama-alpaca}. The Chinese LLaMA model markedly enhances the original LLaMA’s proficiency in understanding and generating Chinese content. 

\subsection{BLOOM}
BLOOM \cite{bloom} is an autoregressive large language model trained on a large corpus of text data with the aid of industrial-scale computational resources. Consequently, it can generate coherent text in 46 languages, including 13 programming languages, which is highly comparable to human-authored text.  Furthermore, BLOOM can be directed to execute text tasks that it hasn't received explicit training for, by framing them as text generation assignments.In our experiments, We use BLOOMZ \cite{bloomz}, a multitask prompted fine-tuned version of BLOOM, with better generalization and zero-shot capabilities.

\subsection{ChatGLM}
ChatGLM is an open-source conversational language model that supports both Chinese and English. It is based on the General Language Model (GLM) \cite{glm,glm130} architecture and consists of 6.2 billion parameters. With the aid of model quantization technology, users can deploy ChatGLM-6B on consumer-grade graphics cards with as little as 6GB of video memory when using the INT4 quantization level. ChatGLM employs a technology similar to ChatGPT, specifically optimized for Chinese question-and-answer (Q\&A) interactions and dialogues. With approximately 1 trillion Chinese-English bilingual tokens, supported by various techniques such as supervision, fine-tuning, and human feedback reinforcement learning, ChatGLM-6B is capable of generating responses that closely align with human preferences.

\subsection{ChatGPT}
ChatGPT is a sibling model to InstructGPT \cite{instrucGPT}, specifically designed to follow user instructions and generate detailed responses. ChatGPT is enhanced through instruction tuning \cite{wei2021finetuned} and reinforcement learning from human feedback (RLHF) \cite{instrucGPT}. Unlike the original GPT-3, which lacks a specific design for following user instructions, ChatGPT demonstrates a significantly improved capability to generate aligned and helpful outputs in response to user instructions. ChatGPT has been widely employed in diverse artificial intelligence scenarios, such as search-based question answering, fundamental NLP tasks, etc.

\section{Experiments and Analyses}
\begin{table*}[!t]
\caption{Experimental results on the FinChina SA dataset. The Longformer model is fine-tuned using the original training set. The large language models in lines two to six are fine-tuned using the single-turn QA form training set. The ChatGPT model predicts the classification result in a zero-shot setting.}
\centering
\label{tab2}
\begin{tabular*}{0.7\hsize}{@{}@{\extracolsep{\fill}}llcccc@{}}
\toprule
&\multirow{2}{*}{Model}&\multicolumn{2}{c}{Sentiment Classification}&\multicolumn{2}{c}{Warning Type}\\
\cmidrule(r){3-4}\cmidrule(r){5-6}
&&Accuracy&Weighted F1&Accuracy&Weighted F1\\
\midrule
1&Longformer&67.45&66.61&54.80&53.76\\
\midrule
2&BLOOMZ&75.52&74.97&66.27&66.29\\
3&Chinese LLaMA&68.67&67.43&55.98&53.63\\
4&Chinese LLaMA Plus&\bf75.99&\bf75.59&\bf67.09&\bf67.13\\
5&ChatGLM(P-tuning v2)&69.48&68.44&52.13&50.75\\
6&ChatGLM&74.76&73.53&65.78&64.08\\
\midrule
7&ChatGPT&46.80&47.43&18.17&18.30\\
\bottomrule
\end{tabular*}
\end{table*}

\begin{table*}[!t]
\caption{Experimental results of models fine-tuned with multi-turn QA form data.}
\centering
\label{tab3}
\begin{tabular*}{0.7\hsize}{@{}@{\extracolsep{\fill}}llcccc@{}}
\toprule
&\multirow{2}{*}{Model}&\multicolumn{2}{c}{Sentiment Classification}&\multicolumn{2}{c}{Warning Type}\\
\cmidrule(r){3-4}\cmidrule(r){5-6}
&&Accuracy&Weighted F1&Accuracy&Weighted F1\\
\midrule
1&BLOOMZ&70.14&72.04&\bf61.69&\bf61.80\\
2&Chinese LLaMA Plus&\bf71.30&\bf73.42&60.05&61.75\\  %70.14 72.04
3&ChatGLM&69.56&70.08&59.40&58.69\\
\bottomrule
\end{tabular*}
\end{table*}

\subsection{Task Definition}
We start by defining the fine-grained financial sentiment analysis task for enterprise early warning, before we introduce our experiments. Given a news text $s$ with a sequence of words $\{w_1,w_2,...w_n\}$ and the names of all companies $\{I_1,I_2,...I_n\}$ involved in this sentence, the goal is to predict their polarities and early warning type.

\subsection{Dataset Construction}

According to two different forms of supervised fine-tuning methods, instruction fine-tuning and dialogue fine-tuning, we construct the dataset into two forms of single-turn QA and multi-turn QA. Specifically, in a single round of QA, the model will be asked to answer the sentiment polarities and warning types of all companies at once. In multiple rounds of QA, the task will be decomposed into multiple subtasks. In each subtask, model will answer the polarity or warning type of a company, and the model will complete all subtasks in turn. The dataset is split into 90\% for training and 10\% for test. 

\subsection{Training Details}
The models are implemented in PyTorch using the Huggingface Transformers package. It should be noted that, in addition to the Longformer model, we choose a version with parameters of about 7B for the open-source large languge models, and there are all links to the original model weight in the footnote\footnote{\url{https://github.com/ymcui/Chinese-LLaMA-Alpaca}}\footnote{\url{https://github.com/THUDM/ChatGLM-6B}}\footnote{\url{https://huggingface.co/bigscience/bloomz-7b1-mt}}. The maximum context length is set to 4,096. The models are trained with the AdamW optimizer, using a batch size of 32 and 3 epochs. The learning rate is set to 2e-5, and weight decay is set to 0. We use technologies such as DeepSpeed ZeRO \cite{zero} and FSDP \cite{FSDP} to optimize the parallel training of models. We assess ChatGPT abilities in a zero-shot manner using our dataset. Figure \ref{FIG_3} shows the illustration of the prompt template used in our experiments for ChatGPT. For the Longformer model, two fully connected layers are connected to perform the classification, and the loss function is cross entropy. 

\subsection{Experimental Results}

Table \ref{tab2} and table \ref{tab3} presents the experimental results on our dataset. To fine-tune the large language models, we preprocess the training set into two forms: single-turn QA and multi-turn QA. The input will consist of task instructions, context, and questions. A single-turn answer should incorporate the polarities and warning type of all companies at once. In the multi-turn QA form, the polarities or warning type of a company are addressed one at a a round of dialogue until all the results are covered.

As shown in table \ref{tab2}, the Longformer model is fine-tuned using the original training set. The models in lines two to six are fine-tuned using the  single-turn QA form training set. The ChatGPT model predicts the classification result using a prompt-based method in a zero-shot setting. The table reveals that Chinese LLaMA Plus outperforms other models in the dataset. The experimental results demonstrate that leveraging a substantial Chinese corpus for secondary pre-training significantly enhances LLaMA's Chinese understanding and generation capabilities. In the sentiment classification task, Chinese LLaMA Plus achieves the highest accuracy of 75.99, followed by BLOOMZ with 75.52 and ChatGLM with 74.46. Another warning type classification task exhibits a similar performance gap among all models. 

It is worth noting that ChatGPT struggles with the financial sentiment analysis task in the zero-shot setting. This result suggests that ChatGPT is not proficient in understanding financial concepts. Evidently, ChatGPT faces challenges in connecting statements to human sentiment in financial news, possibly due to a lack of domain knowledge. Unlike other tasks, where a model can retrieve information from the context and link operations to achieve the final output, the sentiment analysis task demands a deeper understanding of domain-specific expressions and underlying sentiment knowledge. Such a level of understanding presents a challenge for models like ChatGPT, which may have limited exposure to the domain-specific training corpus. The LLMs achieve superior performance compared to the Longformer model after fine-tuning, demonstrating the capability of general generative language models to transfer effectively to specific domains. Moreover, while efficient parameter fine-tuning methods like P-tuning v2 decrease video memory usage, our experiments indicate a significant reduction in the model's accuracy.

In the multi-turn QA form dataset, we adopt a turn-based approach. After each turn, we gather the answer generated by the models, append it to the previous question, and utilize this combined input as the prompt for the subsequent round. The results, as presented in Table \ref{tab3}, indicate that converting single-turn instruction data into a multi-turn QA format and subsequently fine-tuning LLMs may not be beneficial for performance improvement. Through a thorough check of the experimental results, we discovered that training the model with multi-round dialogues increases the likelihood of generating repetitive and erroneous content during the inference phase. One possibility is that fine-tuning models through conversation-based approaches is more appropriate for data that closely resembles real dialogues. In contrast, the transformation of multi-round dialogue data using templates results in a relatively simplistic and less diverse format. We leave it as future work to collect more real financial dialogue data and evaluate more financial tasks.

\section{Conclusion and Future Work}
Our new dataset, FinChina SA, will serve as one of the main directions for the next research focus – how to apply LLMs in the financial domain. We experimented with various large pre-trained language models and found that fine-tuned LLMs perform impressively in sentiment analysis tasks, surpassing the performance of the fine-tuned Longformer model. However, ChatGPT exhibits limitations in this task, which needs handling domain-specific knowledge and terminology. While ChatGPT performs well in generic NLP tasks, its effectiveness in the financial domain is not comparable to that of specialized models fine-tuned specifically for financial tasks. In conclusion, ChatGPT provides a solid foundation for NLP tasks related to finance. However, further improvements can enhance its performance. In addition, transforming single-turn instruction data into multi-turn QA form and then fine-tuning LLMs may be unhelpful to improve the performance. Efficient parameter fine-tuning methods, such as P-tuning v2, reduce memory consumption, but in our experiments significantly reduce the accuracy of the model.

Due to cost constraints of ChatGPT API and training LLMs, we have only tested open-source LLMs with about 7B parameters on relatively small data, and we do not conduct extensive experiments on complex prompt engineering for ChatGPT. We believe our experiments can provide valuable insights into the task of sentiment analysis over real-world specific domains and facilitate further improvements. Meanwhile, we do not exclude the possibility that there could be better performances for prompting-based methods if applying advanced prompt engineering or GPT-4, which is costlier. We leave this for future work. We plan to try more open-source LLMs with larger parameters and release a larger financial domain dataset in the future.

\bibliographystyle{named}
\bibliography{ijcai23}

\begin{thebibliography}{}

\bibitem[\protect\citeauthoryear{Beltagy \bgroup \em et al.\egroup
  }{2020}]{Beltagy2020Longformer}
Iz~Beltagy, Matthew~E. Peters, and Arman Cohan.
\newblock Longformer: The long-document transformer.
\newblock {\em arXiv:2004.05150}, 2020.

\bibitem[\protect\citeauthoryear{Brown \bgroup \em et al.\egroup }{2020}]{gpt3}
Tom Brown, Benjamin Mann, Nick Ryder, Melanie Subbiah, Jared~D Kaplan, Prafulla
  Dhariwal, Arvind Neelakantan, Pranav Shyam, Girish Sastry, Amanda Askell,
  et~al.
\newblock Language models are few-shot learners.
\newblock {\em Advances in neural information processing systems},
  33:1877--1901, 2020.

\bibitem[\protect\citeauthoryear{Chen \bgroup \em et al.\egroup
  }{2023}]{chen2023phoenix}
Zhihong Chen, Feng Jiang, Junying Chen, Tiannan Wang, Fei Yu, Guiming Chen,
  Hongbo Zhang, Juhao Liang, Chen Zhang, Zhiyi Zhang, et~al.
\newblock Phoenix: Democratizing chatgpt across languages.
\newblock {\em arXiv preprint arXiv:2304.10453}, 2023.

\bibitem[\protect\citeauthoryear{Chiang \bgroup \em et al.\egroup
  }{2023}]{vicuna2023}
Wei-Lin Chiang, Zhuohan Li, Zi~Lin, Ying Sheng, Zhanghao Wu, Hao Zhang, Lianmin
  Zheng, Siyuan Zhuang, Yonghao Zhuang, Joseph~E. Gonzalez, Ion Stoica, and
  Eric~P. Xing.
\newblock Vicuna: An open-source chatbot impressing gpt-4 with 90\%* chatgpt
  quality, March 2023.

\bibitem[\protect\citeauthoryear{Cortis \bgroup \em et al.\egroup
  }{2017}]{Semeval2017}
Keith Cortis, Andr{\'e} Freitas, Tobias Daudert, Manuela Huerlimann, Manel
  Zarrouk, Siegfried Handschuh, and Brian Davis.
\newblock Semeval-2017 task 5: Fine-grained sentiment analysis on financial
  microblogs and news.
\newblock In {\em Proceedings of the 11th international workshop on semantic
  evaluation (SemEval-2017)}, pages 519--535, 2017.

\bibitem[\protect\citeauthoryear{Cui \bgroup \em et al.\egroup
  }{2023}]{chinese-llama-alpaca}
Yiming Cui, Ziqing Yang, and Xin Yao.
\newblock Efficient and effective text encoding for chinese llama and alpaca.
\newblock {\em arXiv preprint arXiv:2304.08177}, 2023.

\bibitem[\protect\citeauthoryear{Devlin \bgroup \em et al.\egroup
  }{2018}]{devlin2018bert}
Jacob Devlin, Ming-Wei Chang, Kenton Lee, and Kristina Toutanova.
\newblock Bert: Pre-training of deep bidirectional transformers for language
  understanding.
\newblock {\em arXiv preprint arXiv:1810.04805}, 2018.

\bibitem[\protect\citeauthoryear{Do \bgroup \em et al.\egroup
  }{2019}]{do2019deep}
Hai~Ha Do, Penatiyana~WC Prasad, Angelika Maag, and Abeer Alsadoon.
\newblock Deep learning for aspect-based sentiment analysis: a comparative
  review.
\newblock {\em Expert systems with applications}, 118:272--299, 2019.

\bibitem[\protect\citeauthoryear{Du \bgroup \em et al.\egroup }{2022}]{glm}
Zhengxiao Du, Yujie Qian, Xiao Liu, Ming Ding, Jiezhong Qiu, Zhilin Yang, and
  Jie Tang.
\newblock Glm: General language model pretraining with autoregressive blank
  infilling.
\newblock In {\em Proceedings of the 60th Annual Meeting of the Association for
  Computational Linguistics (Volume 1: Long Papers)}, pages 320--335, 2022.

\bibitem[\protect\citeauthoryear{Li \bgroup \em et al.\egroup
  }{2023}]{li2023chatgpt}
Xianzhi Li, Xiaodan Zhu, Zhiqiang Ma, Xiaomo Liu, and Sameena Shah.
\newblock Are chatgpt and gpt-4 general-purpose solvers for financial text
  analytics? an examination on several typical tasks.
\newblock {\em arXiv preprint arXiv:2305.05862}, 2023.

\bibitem[\protect\citeauthoryear{Maia \bgroup \em et al.\egroup
  }{2018}]{www2018open}
Macedo Maia, Siegfried Handschuh, Andr{\'e} Freitas, Brian Davis, Ross
  McDermott, Manel Zarrouk, and Alexandra Balahur.
\newblock Www'18 open challenge: financial opinion mining and question
  answering.
\newblock In {\em Companion proceedings of the the web conference 2018}, pages
  1941--1942, 2018.

\bibitem[\protect\citeauthoryear{Muennighoff \bgroup \em et al.\egroup
  }{2022}]{bloomz}
Niklas Muennighoff, Thomas Wang, Lintang Sutawika, Adam Roberts, Stella
  Biderman, Teven~Le Scao, M~Saiful Bari, Sheng Shen, Zheng-Xin Yong, Hailey
  Schoelkopf, et~al.
\newblock Crosslingual generalization through multitask finetuning.
\newblock {\em arXiv preprint arXiv:2211.01786}, 2022.

\bibitem[\protect\citeauthoryear{Ouyang \bgroup \em et al.\egroup
  }{2022}]{instrucGPT}
Long Ouyang, Jeffrey Wu, Xu~Jiang, Diogo Almeida, Carroll Wainwright, Pamela
  Mishkin, Chong Zhang, Sandhini Agarwal, Katarina Slama, Alex Ray, et~al.
\newblock Training language models to follow instructions with human feedback.
\newblock {\em Advances in Neural Information Processing Systems},
  35:27730--27744, 2022.

\bibitem[\protect\citeauthoryear{Pang \bgroup \em et al.\egroup
  }{2008}]{pang2008opinion}
Bo~Pang, Lillian Lee, et~al.
\newblock Opinion mining and sentiment analysis.
\newblock {\em Foundations and Trends{\textregistered} in information
  retrieval}, 2(1--2):1--135, 2008.

\bibitem[\protect\citeauthoryear{Radford \bgroup \em et al.\egroup
  }{2018}]{gpt1}
Alec Radford, Karthik Narasimhan, Tim Salimans, Ilya Sutskever, et~al.
\newblock Improving language understanding by generative pre-training.
\newblock 2018.

\bibitem[\protect\citeauthoryear{Radford \bgroup \em et al.\egroup
  }{2019}]{gpt2}
Alec Radford, Jeffrey Wu, Rewon Child, David Luan, Dario Amodei, Ilya
  Sutskever, et~al.
\newblock Language models are unsupervised multitask learners.
\newblock {\em OpenAI blog}, 1(8):9, 2019.

\bibitem[\protect\citeauthoryear{Raffel \bgroup \em et al.\egroup }{2020}]{T5}
Colin Raffel, Noam Shazeer, Adam Roberts, Katherine Lee, Sharan Narang, Michael
  Matena, Yanqi Zhou, Wei Li, and Peter~J Liu.
\newblock Exploring the limits of transfer learning with a unified text-to-text
  transformer.
\newblock {\em The Journal of Machine Learning Research}, 21(1):5485--5551,
  2020.

\bibitem[\protect\citeauthoryear{Rajbhandari \bgroup \em et al.\egroup
  }{2020}]{zero}
Samyam Rajbhandari, Jeff Rasley, Olatunji Ruwase, and Yuxiong He.
\newblock Zero: Memory optimizations toward training trillion parameter models.
\newblock In {\em SC20: International Conference for High Performance
  Computing, Networking, Storage and Analysis}, pages 1--16. IEEE, 2020.

\bibitem[\protect\citeauthoryear{Scao \bgroup \em et al.\egroup }{2022}]{bloom}
Teven~Le Scao, Angela Fan, Christopher Akiki, Ellie Pavlick, Suzana Ili{\'c},
  Daniel Hesslow, Roman Castagn{\'e}, Alexandra~Sasha Luccioni, Fran{\c{c}}ois
  Yvon, Matthias Gall{\'e}, et~al.
\newblock Bloom: A 176b-parameter open-access multilingual language model.
\newblock {\em arXiv preprint arXiv:2211.05100}, 2022.

\bibitem[\protect\citeauthoryear{Taori \bgroup \em et al.\egroup
  }{2023}]{alpaca}
Rohan Taori, Ishaan Gulrajani, Tianyi Zhang, Yann Dubois, Xuechen Li, Carlos
  Guestrin, Percy Liang, and Tatsunori~B. Hashimoto.
\newblock Stanford alpaca: An instruction-following llama model.
\newblock \url{https://github.com/tatsu-lab/stanford_alpaca}, 2023.

\bibitem[\protect\citeauthoryear{Touvron \bgroup \em et al.\egroup
  }{2023}]{llama}
Hugo Touvron, Thibaut Lavril, Gautier Izacard, Xavier Martinet, Marie-Anne
  Lachaux, Timoth{\'e}e Lacroix, Baptiste Rozi{\`e}re, Naman Goyal, Eric
  Hambro, Faisal Azhar, et~al.
\newblock Llama: Open and efficient foundation language models.
\newblock {\em arXiv preprint arXiv:2302.13971}, 2023.

\bibitem[\protect\citeauthoryear{Vaswani \bgroup \em et al.\egroup
  }{2017}]{transformer}
Ashish Vaswani, Noam Shazeer, Niki Parmar, Jakob Uszkoreit, Llion Jones,
  Aidan~N Gomez, {\L}ukasz Kaiser, and Illia Polosukhin.
\newblock Attention is all you need.
\newblock {\em Advances in neural information processing systems}, 30, 2017.

\bibitem[\protect\citeauthoryear{Wang \bgroup \em et al.\egroup
  }{2022}]{self-instruct}
Yizhong Wang, Yeganeh Kordi, Swaroop Mishra, Alisa Liu, Noah~A Smith, Daniel
  Khashabi, and Hannaneh Hajishirzi.
\newblock Self-instruct: Aligning language model with self generated
  instructions.
\newblock {\em arXiv preprint arXiv:2212.10560}, 2022.

\bibitem[\protect\citeauthoryear{Wei \bgroup \em et al.\egroup
  }{2021}]{wei2021finetuned}
Jason Wei, Maarten Bosma, Vincent~Y Zhao, Kelvin Guu, Adams~Wei Yu, Brian
  Lester, Nan Du, Andrew~M Dai, and Quoc~V Le.
\newblock Finetuned language models are zero-shot learners.
\newblock {\em arXiv preprint arXiv:2109.01652}, 2021.

\bibitem[\protect\citeauthoryear{Wu \bgroup \em et al.\egroup
  }{2023}]{wu2023bloomberggpt}
Shijie Wu, Ozan Irsoy, Steven Lu, Vadim Dabravolski, Mark Dredze, Sebastian
  Gehrmann, Prabhanjan Kambadur, David Rosenberg, and Gideon Mann.
\newblock Bloomberggpt: A large language model for finance.
\newblock {\em arXiv preprint arXiv:2303.17564}, 2023.

\bibitem[\protect\citeauthoryear{Yang \bgroup \em et al.\egroup
  }{2023}]{yang2023fingpt}
Hongyang Yang, Xiao-Yang Liu, and Christina~Dan Wang.
\newblock Fingpt: Open-source financial large language models.
\newblock {\em arXiv preprint arXiv:2306.06031}, 2023.

\bibitem[\protect\citeauthoryear{Zeng \bgroup \em et al.\egroup
  }{2022}]{glm130}
Aohan Zeng, Xiao Liu, Zhengxiao Du, Zihan Wang, Hanyu Lai, Ming Ding, Zhuoyi
  Yang, Yifan Xu, Wendi Zheng, Xiao Xia, et~al.
\newblock Glm-130b: An open bilingual pre-trained model.
\newblock {\em arXiv preprint arXiv:2210.02414}, 2022.

\bibitem[\protect\citeauthoryear{Zhao \bgroup \em et al.\egroup }{2023}]{FSDP}
Yanli Zhao, Andrew Gu, Rohan Varma, Liang Luo, Chien-Chin Huang, Min Xu, Less
  Wright, Hamid Shojanazeri, Myle Ott, Sam Shleifer, et~al.
\newblock Pytorch fsdp: Experiences on scaling fully sharded data parallel.
\newblock {\em arXiv preprint arXiv:2304.11277}, 2023.

\end{thebibliography}
\end{document}